\documentclass[11pt]{article}

\newcommand{\keywords}[1]{%
  \begin{center}
    \vspace{0.5em}
    \textbf{Keywords:} #1
  \end{center}
}

\usepackage[square,sort,comma,numbers]{natbib}

\usepackage{hyperref}  % Hyperref should always be loaded last
\usepackage{fullpage}         % Set standard margins
\usepackage{amsmath, amssymb} % Math packages
\usepackage{graphicx}         % Include images

\usepackage{amsmath}
\usepackage{graphicx}
\usepackage{booktabs}
\usepackage{multirow}

\usepackage{hyperref}
\usepackage{caption}
\usepackage{subcaption}

\usepackage[T1]{fontenc}
\begin{document}
\newcommand{\xmark}{\ding{55}}
\newcommand{\tick}{\ding{51}}
\title{VibrantVS: A High-Resolution Multi-Task Transformer for Forest Canopy Height Estimation}

%
%
% \titlerunning (removed for article class){VibrantVS}

% Authors, for the paper (add full first names)

\author{Tony Chang $^{1}$, Kiarie Ndegwa $^{1}$, Andreas Gros $^{1}$\thanks{Corresponding author: \textbf{andi@vibrantplanet.net}},\\ Vincent A. Landau $^{1}$, Luke J. Zachmann $^{1}$, Bogdan State $^{1}$, Mitchell A. Gritts $^{1}$,\\ Colton W. Miller $^{1}$, Nathan E. Rutenbeck $^{1}$, Scott Conway $^{1}$, and Guy Bayes $^{1}$}

\date{
\vspace{1em}
{$^{1}$ Vibrant Planet Public Benefit Corporation\\ 
11025 Pioneer Trail Unit 206 Truckee,CA, USA\\
}
\vspace{1em}
}

\maketitle

\begin{abstract}
This paper explores the application of a novel multi-task vision transformer (ViT) model for the estimation of canopy height models (CHMs) using 4-band National Agriculture Imagery Program (NAIP) imagery across the western United States. We compare the effectiveness of this model in terms of accuracy and precision aggregated across ecoregions and class heights versus three other benchmark peer-reviewed models. Key findings suggest that, while other benchmark models can provide high precision in localized areas, the VibrantVS model has substantial advantages across a broad reach of ecoregions in the western United States with higher accuracy, higher precision, the ability to generate updated inference at a cadence of three years or less, and high spatial resolution. The VibrantVS model provides significant value for ecological monitoring and land management decisions, including for wildfire mitigation.

\end{abstract}

\section{Introduction}
\label{sec:intro}
During the past three decades, increases in the number of wildfires and burned areas across the western United States have negatively impacted human health, assets, and ecological structure and function \citep{moritz2014learning, westerling_increasing_2016}. This fire activity, compounded by recent widespread drought and historic fire suppression \citep{abatzoglou2016impact}, has led to forest conditions that lack sufficient resilience to recover from natural disturbances and thereby threatens the ecosystem services on which human communities depend \citep{stevens2018evidence, hessburg2005dry}. Such conditions have required forest managers to prioritize the restoration of historical fire regimes and the treatment and maintenance of forests to reduce the potential for high-severity fire \citep{hoffman2021conservation}. A major requirement in the identification of appropriate treatments and restoration actions is access to current spatial data that accurately describes the horizontal and vertical structure of vegetation. 

Canopy height models (CHMs) describe the height of the canopy top from the ground. The availability of current canopy height data in a three-dimensional georeferenced surface \citep{van2010retrieval} is crucial for various ecological applications including biomass estimation \citep{means1999use}, predicting fire behavior \citep{Riano2003, andersen2005estimating} individual tree detection \citep{silva2016imputation, popescu2004seeing}, and habitat quality assessment \citep{vierling2008lidar}. Accurate quantification of canopy height not only enhances our understanding of forest dynamics \citep{Lefsky2002, beland2019promoting, wulder2008role}, but also facilitates strategic, tactical, and operational planning decisions to improve forest health, promote biodiversity, and protect threatened and endangered species \citep{white2016remote}. Therefore, the development of a high resolution continuous CHM across large (in excess of 1 million ha), multi-jurisdictional landscapes such as the western United States is critical for the successful restoration, monitoring, and management of forests. 

Lidar data has emerged as the gold standard for three-dimensional mapping of vegetation structure due to its high spatial fidelity and ability to penetrate forest canopies \citep{olszewski2022lidar, kramer2014quantifying, Lefsky2002}. However, even with economies of scale \citep{franklin2002change} and cost effectiveness compared to conducting field inventories over large areas \citep{means1999use}, lidar datasets are constrained by their limited spatial and temporal coverage. These constraints limit the ability of lidar acquisitions to reflect up-to-date forest structure because of the dynamic nature of forest ecosystems as they respond to disturbances, treatments, natural growth, or climate change \citep{dubayah2000lidar}. On the other hand, the proliferation of freely available, remotely sensed data with consistent temporal revisit periods, such as the satellite-based imagery of Sentinel-2 \citep{drusch2012sentinel}, or aerial imagery from NAIP, has opened new avenues to methodologies that can be tailored to model vegetation structure with high spatial and temporal resolution over longer periods of time and over larger areas \citep{valbuena2020standardizing, WAGNER2024114099}. 

The availability of large remotely sensed image libraries in conjunction with newer deep learning methods has allowed for the emergence of a suite of modeled canopy height products providing continuous high resolution (1m-30m) forest structure measurements across broad spatial extents \citep{tolan2024very, lang2023high, li2023deep}. This paper aims to benchmark a selection of the latest peer-reviewed deep learning/machine learning-based CHMs across varied ecosystems in the western United States. Additionally, we describe our own multi-task vision transformer model based on NAIP imagery: VibrantVS. We present an analysis that will: 1) address which model is optimal based on standard machine learning metrics compared to aerial lidar measurements to serve as a benchmark for the broad remote sensing community; and 2) compare model estimates across a variety of ecoregions in the western United States to better understand the utility of these models for fine-grained ecological management and decision making.

\section{Study Area and Data}
\begin{figure}[htbp!]
    \centering
    \includegraphics[width=0.5\textwidth]{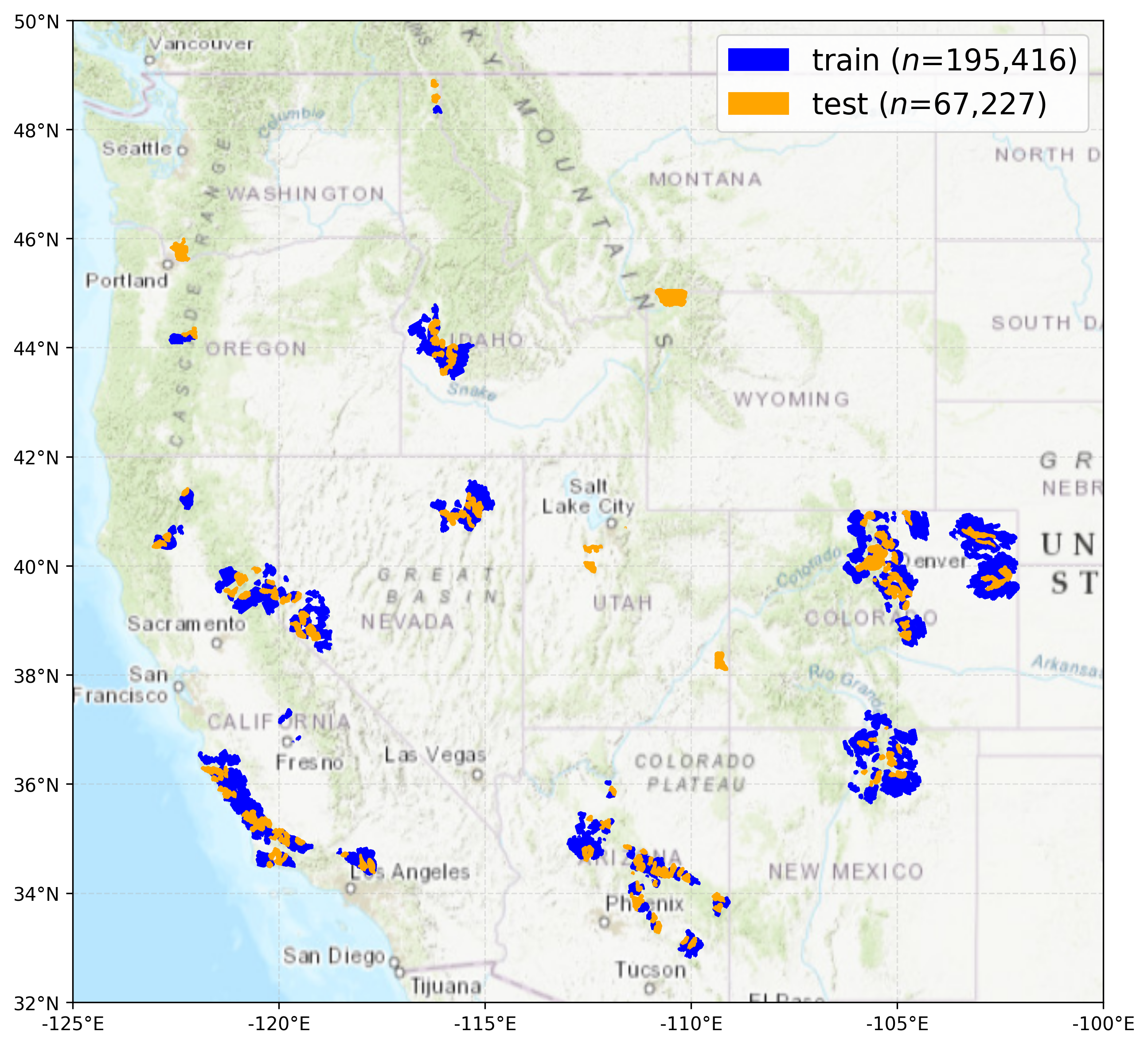}  
    \caption{Sampling of tiles within Hydrologic Units 12 (HUC12) watersheds of the western United States that covers 24 EPA L3 ecoregions containing sufficient quality 3DEP lidar data and spatially/temporally intersecting NAIP data. Regions were additionally selected within WCS areas to optimize for model evaluation in regions where wildfire risk mitigation is a priority.}
    \label{fig:study_area}  
\end{figure}

In this study, we compare CHMs for the western contiguous United States.  Given our focus on forest structure data that would have the greatest utility in understanding wildfire impacts, we focus our sampling to intersect (Fig. \ref{fig:study_area}): 
\begin{enumerate}
    \item The Wildfire Crisis Strategy (WCS) landscapes defined by the United States Forest Service (USFS) to prioritize regions at the greatest risk of severe wildfire. 
    \citep{us_forest_service_2022}.
    \item National Ecological Observatory Network (NEON) research sites within the western United States, due to their independent collection of individual field-based tree height measurements and aerial lidar against which to validate model predictions \citep{thibault2023us}.
    \item Areas that intersect the spatial extent for the 3D Elevation Program (3DEP) work units in the Work Unit Extent Spatial Metadata (WESM) dataset \citep{wesm} that meet the seamless and 1-meter DEM quality criteria. 
\end{enumerate}

To quantify the ability of the models to robustly predict forest canopy height across an ecological gradient, we stratify our data by Environmental Protection Agency (EPA) Level 3 (L3) ecoregions \citep{omernik2014ecoregions}. EPA L3 ecoregions are delineated based on relatively homogeneous environmental conditions, such as vegetation types, soil, climate, and landforms. This coherence reduces variability in environmental factors unrelated to canopy height within each ecoregion, making it easier to evaluate the performance of the baseline models across a well-defined set of environmental conditions. Additionally, EPA L3 ecoregions provide a manageable range of dominant species and climate variability while still capturing key regional distinctions.
Finally, evaluating CHM performance at this ecoregion scale can directly inform practices in forest management, carbon accounting, and biodiversity conservation.

\begin{figure*}[htbp!]
    \centering
    \includegraphics[width=\textwidth]{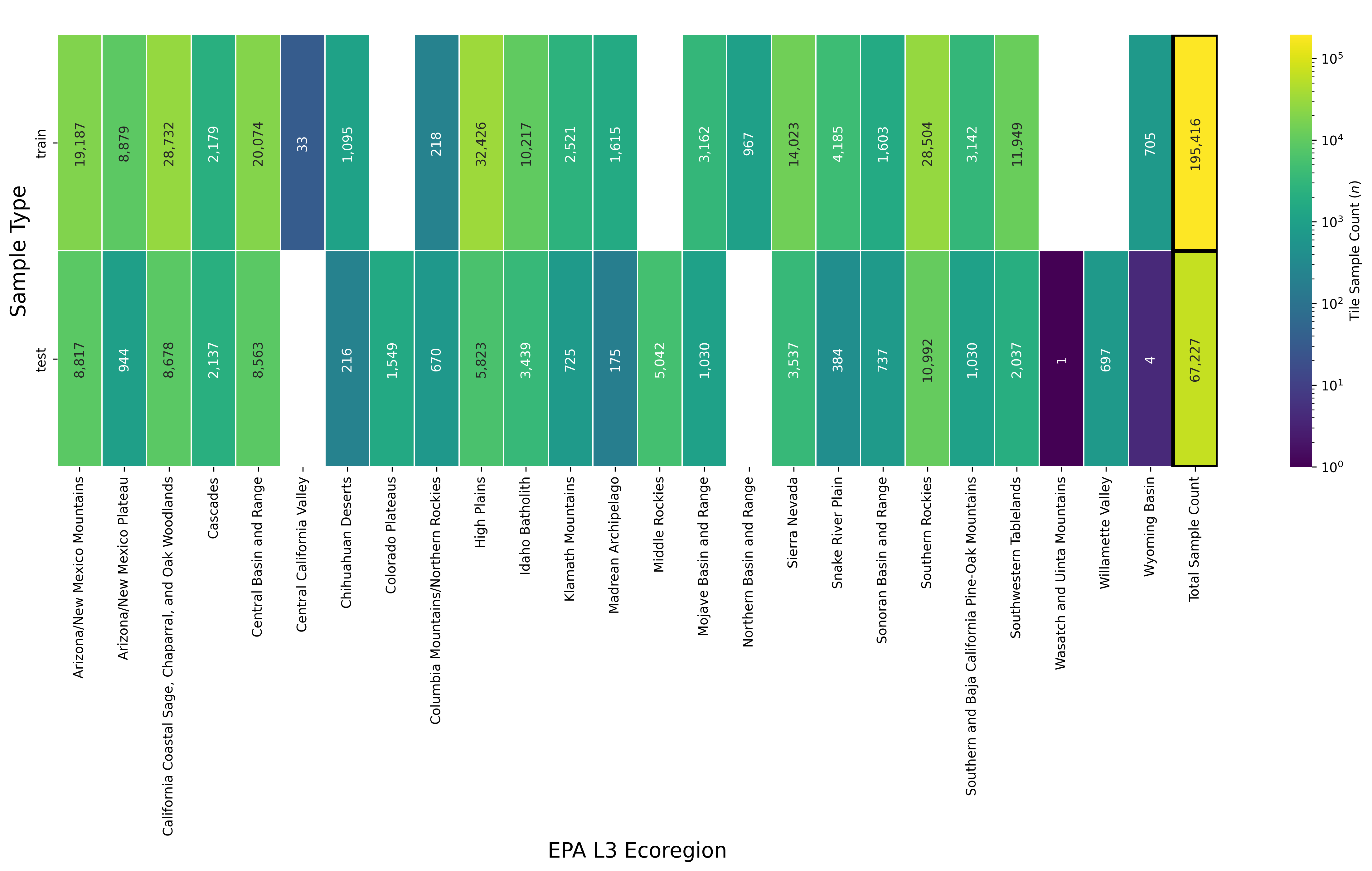}  
    \caption{Sample tile counts within each of the randomly sampled train and test (approx. 85\% to 15\% ratio) groups by EPA L3 ecoregion.}
    \label{fig:study_area_counts}  
\end{figure*}

We generated a total of 262,643 sample tiles with a spatial footprint of 0.5 $\times$ 0.5 km\textsuperscript{2} (1,000 $\times$ 1,000 pixel rasters at 0.5-meter resolution) that crossed a total of 24 different EPA L3 ecoregions (Fig.\ref{fig:study_area_counts}), providing us with a robust representation of tree vegetation data and lidar CHM heights (Fig.\ref{app:ecoregion_histograms}). We split the data into two groups: 1) a `train' set of 195,416
samples, which contains data for training and validation of the VibrantVS ViT model; and 2) a test set of 
67,227 samples, which contain tiles withheld from the model to evaluate performance. The lidar collection dates range from 2014 to 2021 (Table \ref{tab:lidar_summary}).

\begin{figure*}[htbp!]
    \centering
    \includegraphics[width=\textwidth]{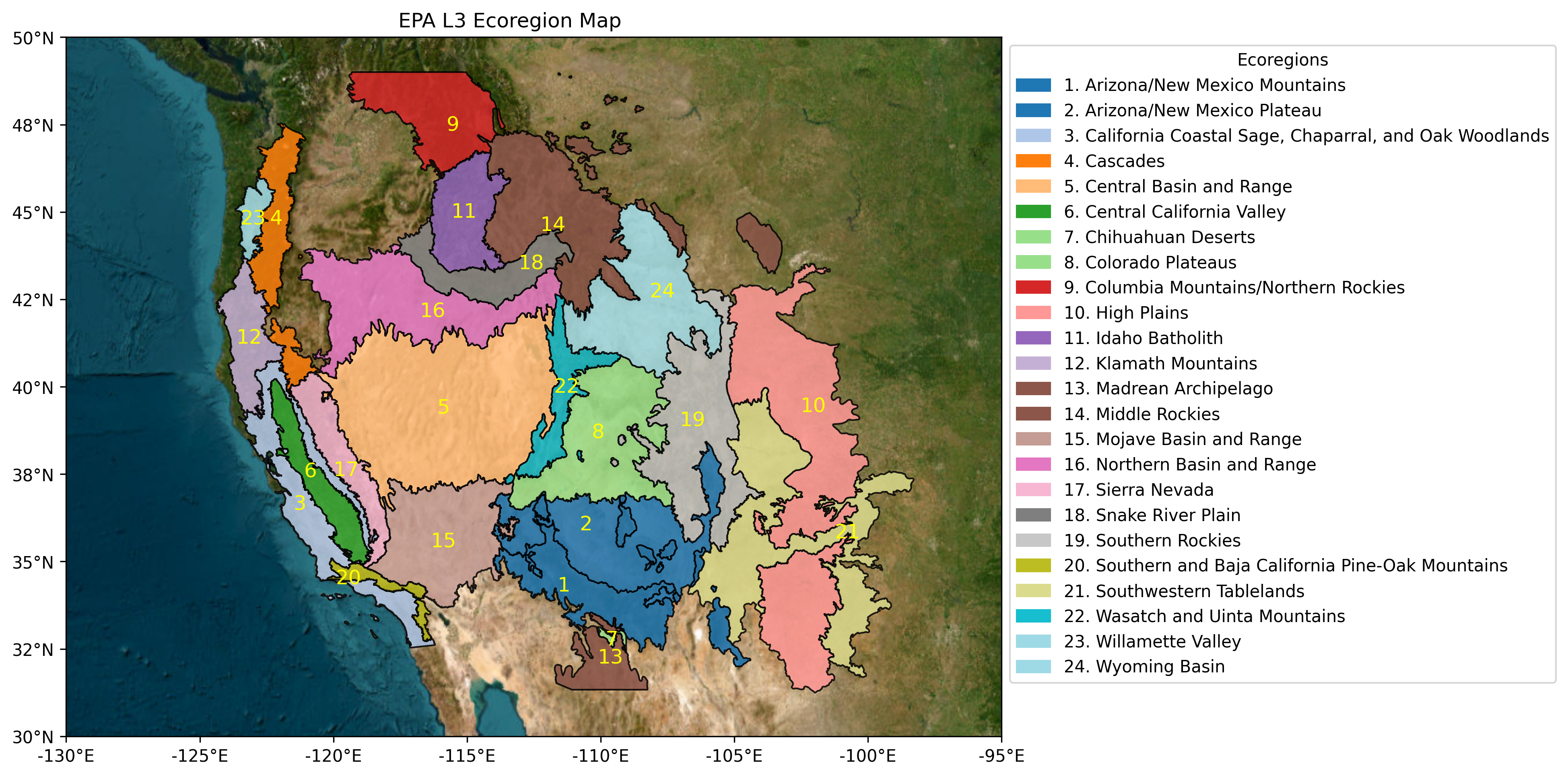}  
    \caption{Map of EPA L3 Ecoregions  into which our test tiles were aggregated to evaluate baseline model performance.}
    \label{fig:ecoregion_map}  e
\end{figure*}

\begin{table}[ht]
    \centering
    
    \caption{Summary of samples by lidar collection year and unique EPA L3 ecoregions. Sample tiles are 1 $\times$ 1 km$^2$, so $n$ samples is also the total area in ($km^2$).}
    \begin{tabular}{l|r|r}
    \toprule
    \textbf{Train}  \\
    \hline
    Lidar acq. year & $n$ samples & \# unique ecoregions \\
    \midrule
    2014 & 523 & 2 \\
    2015 & 6,568 & 2 \\
    2016 & 24,838 & 7 \\
    2017 & 32,880 & 8 \\
    2018 & 54,114 & 11 \\
    2019 & 24,974 & 11 \\
    2020 & 44,003 & 6 \\
    2021 & 7,516 & 6 \\
    \hline
    \hline
    \textbf{Test}  \\
    \hline
    Lidar acq. year & $n$ samples & \# unique ecoregions \\
    \hline
    2015 & 1,384 & 2 \\
    2016 & 7,966 & 6 \\
    2017 & 9,616 & 7 \\
    2018 & 13,747 & 12 \\
    2019 & 8,748 & 9 \\
    2020 & 22,459 & 7 \\
    2021 & 3,307 & 6 \\
    \bottomrule
    \end{tabular}
    \label{tab:lidar_summary}
\end{table}

\subsection{Predictor Data}
\subsubsection{NAIP Imagery}
We utilized 4-band NAIP imagery, which includes RGB and near-infrared (NIR) bands \citep{earth_resources_observation_and_science_eros_center_national_2017}. NAIP data are collected at the state level in varying temporal frequencies (originally at five-year cycles and every three years or less since 2009) and at varying spatial resolutions from 0.3 meters to 1 meter since the year 2002 to the present. To create uniform training, test, and inference datasets we collected data from the AWS Open Data Registry \citep{aws_naip} and resampled the NAIP data to a common spatial resolution of 0.5 meters using a bilinear interpolation for the years of 2014-2022. We tiled the NAIP data into 1 $\times$ 1 km\textsuperscript{2} spatial footprint tiles and stored them within AWS S3 storage buckets in Cloud-Optimized GeoTIFF format. We filtered NAIP tile data to be within one year of the spatially coincident lidar acquisition date to reduce opportunities for incorrect labels for the same tile due to disturbances such as wildfires. 

\subsection{Label Data}
\subsubsection{3DEP Lidar}
We downloaded available lidar data developed by the USGS 3DEP program \citep{sugarbaker20143d}. All lidar geospatial metadata is available by work unit in the Work Unit Extent Spatial Metadata (WESM) GeoPackage that is published daily \citep{wesm}. The WESM contains information about lidar point clouds and which source digital elevation model (DEM) work units have been processed and made available to the public. It includes metadata about work units, including quality level, data acquisition dates, and links to project-level metadata. We developed automated scripts to download raw `laz' files from the associated download URLs. We used the `lidR' package in R \citep{lidR} to process the data as follows:
\begin{enumerate}
\item{download, reproject, tile raw lidar data}
\item{create outlier filtered normalized point clouds}
\item{create Digital Terrain Models (DTM)}
\item{create Digital Surface Models (DSM)} 
\item{create Canopy Height Models (CHM)}
\end{enumerate}

These processed lidar assets are stored in 0.5 $\times$ 0.5 km\textsuperscript{2} tiles within our AWS S3 storage buckets in Cloud-Optimized GeoTIFF format at a 0.5-meter pixel size (1,000 $\times$ 1,000 pixels). 

\subsection{Baseline Evaluation Data}
In this study we evaluate data from three existing forest canopy height models as a baseline in comparison to the VibrantVS model.  In order to have greater uniformity in the data, all baseline model outputs were resampled to a 0.5-meter resolution with nearest neighbor resampling and tiled into 1 $\times$ 1km$^2$ spatial footprint tiles.

\subsubsection{Meta Data For Good (Meta) High Resolution Canopy Height - DINOv2 Architecture}
To assess the benefit of the self-supervised training phase on satellite data, we use output data from Meta's state-of-the-art DINOv2 vision encoder as baseline CHM data \citep{tolan2024very}. Meta's approach consists of three different models trained in separate stages, namely an encoder, a dense prediction transformer, and a correction and rescaling network. Four different datasets are used: Maxar Vivid2 0.5-meter resolution mosaics \citep{maxar_vivid2_2024}, NEON aerial lidar surveyed CHMs, Global Ecosystem Dynamics Investigation (GEDI) data and a labeled 9000 tile tree/no tree segmentation dataset \citep{dubayah2022gedi}. At the time of writing this manuscript, this model was one of the only global CHMs with documented performance at 1-meter spatial resolution for the year representing 2020. We downloaded these data provided by the World Resources Institute, which are stored on AWS \citep{meta_wri_chm_2024}, for analysis.

\subsubsection{LANDFIRE Forest Canopy Height model}
The Landscape Fire and Resource Management Planning Tools (LANDFIRE) Project is a program developed to map the characteristics of wildland fuels, vegetation, and fire regime in the United States since 2001 \citep{rollins2006overview}. Due to the national scope and complete spatial coverage across the entire United States, the LANDFIRE fuels dataset is a standard data input in North American fire models \citep{Watts2017,Steenburgh2012,Lane2011}. In this study, we compare the LANDFIRE Forest Canopy Height (FCH) dataset with lidar estimates. The LANDFIRE FCH layer uses regression-tree-based methods to model the relationships between field-measured height with spectral information from Landsat, as well as landscape feature information such as topography and biophysical gradient layers \citep{rollins2009landfire}. We collected this data from the LANDFIRE data portal for the year 2020 \citep{LANDFIREFuels2022}.

\subsubsection{ETH Global canopy height model}
The Lang et al. (\citeyear{lang2023high}) multi-modal model is a deep learning ensemble. This ensemble was trained using a multi-modal set of data, namely 10-meter resolution Sentinel-2-L2A multi-spectral images, sine-cosine embeddings of longitudinal coordinates, and sparse GEDI lidar data. These input data are used to model canopy top height and variance.

Model inference in this ETH model represents the year 2020 and all tiles were downloaded from the ETH Zurich library catalog and repository system \citep{ethz_2023}.

\section{Methodology}
\subsection{Vibrant Planet multi-task vegetation structure ViT: VibrantVS}
\begin{figure*}[htbp!]
    \centering    
    \includegraphics[width=\textwidth]{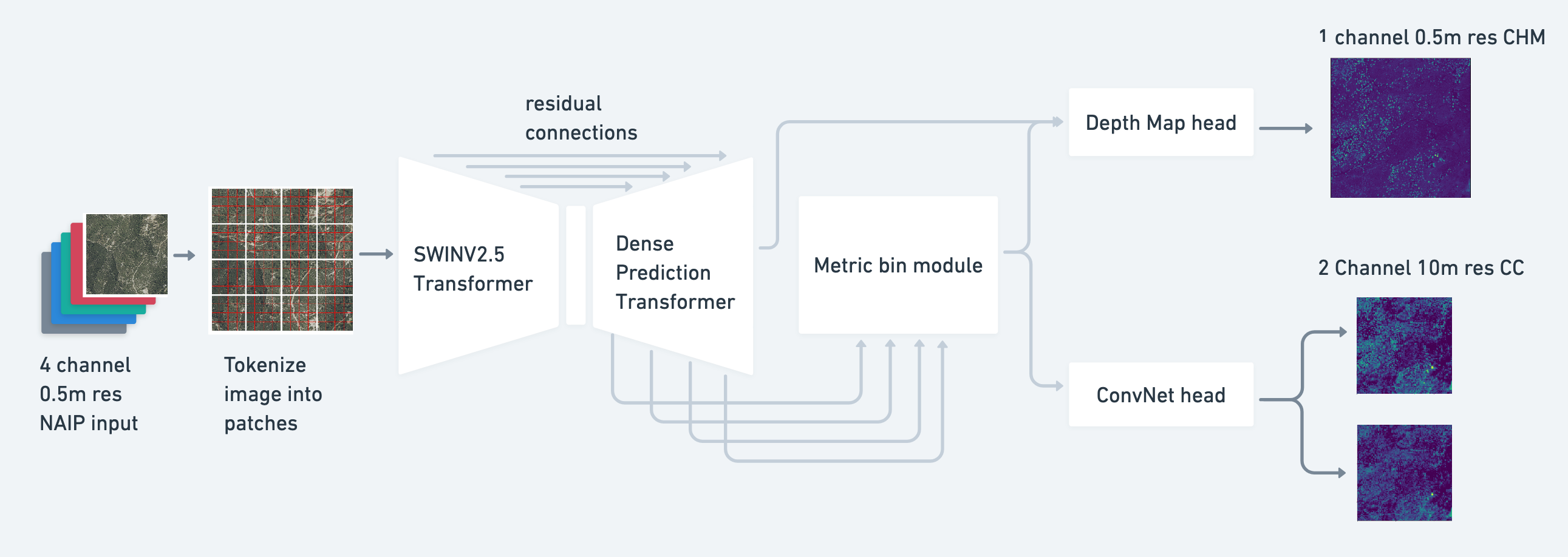}  
    \caption{VibrantVS multi-task vision transformer architecture with 4 band NAIP input to predict CHM and CC.}    
    \label{fig:vibrantvs_model}  
\end{figure*}

The VibrantVS model consists of three main components: a base model and two heads. The base model employs an encoder-decoder Vision Transformer (ViT) architecture, while the heads are a metric-bin module head \citep{bhat2023zoedepth} and a light-weight convolution head. 
The encoder is derived from a modified version of the SWINv2 model \citep{liu2022swin}, while the decoder uses a dense prediction transformer (DPT) \citep{ranftl2020towards} (Fig. \ref{fig:vibrantvs_model}). The DPT module processes features generated by the encoder to produce a normalized relative depth map. These features, along with dense features from the encoder via skip connections, are subsequently input into the metric-bin module head to predict canopy height and into the convolutional prediction head to predict canopy cover (CC). 

The metric-bin module head learns the height distribution for each pixel, represented by 64 bins. These bins are linearly combined to generate dense features that are subsequently merged with feature tensors from the encoder through residual layers. The final output provides tree height metrics at a resolution of 0.5 meters and canopy cover at a resolution of 10 meters. Note that although this model produces both CHM and CC, we only evaluated the CHM portion in this research.

VibrantVS uses a custom modified version of SWINv2 to incorporate the latest transformer-architecture advancements, including token registers, as described in \citep{darcet2023vision}, this stabilized training and reduced noise in latent-attention features. Furthermore, we reconfigured the attention module to employ grouped query attention \citep{ainslie2023gqa}, Flash Attention V2 \citep{dao2023flashattention}, and SWIGLU \citep{shazeer2020glu} activation with RMSNORM \citep{zhang2019root}. These modifications result in reduced training time and improved model convergence stability.

Certain recent optimization techniques, such as schedule-free optimization \citep{defazio2024road}, did not converge effectively. However, the 8-bit Adam optimizer \citep{DBLP:journals/corr/abs-2110-02861} demonstrated satisfactory performance and accelerated training compared to the traditional float32-based AdamW optimizer.

We extended the inference context window from 384x384 pixels to 1,536x1,536 pixels, enabling the model to produce more consistent and artifact-free outputs when handling NAIP inputs of varying quality, including optical artifacts such as seam lines resulting from the merging of data from different flight paths.

The model was trained end to end in a multi-task learning setting to predict both canopy height and canopy cover. Empirical results indicated that L1 loss outperformed Sigloss, providing improved training stability while requiring fewer learnable hyperparameters. The model was trained using input NAIP tiles, paired with dense canopy height data at a 0.5-meter resolution and canopy cover data at a 10-meter resolution. The dataset, approximately 10 terabytes in size, required 2,688 hours of training on an Nvidia 80GB A100 GPU.

To mitigate tiling checkerboard artifacts during inference, the `PyTorch' functions `fold' and `unfold' were replaced with an in-memory buffer cache that makes use of memory pointers. This optimization facilitates the seamless merging of multiple inferences across tiles much larger than the model's context window, enabling inference over regions up to 100,000 $\times$ 100,000 pixels using a single Nvidia 24GB A10G GPU.

We summarize VibrantVS and the baseline models  (Meta, LANDFIRE, and ETH) in Table \ref{app:chm_models}.

\subsection{Calculating error metrics}
We applied a number of error metrics to all baseline models to evaluate the performance of the CHMs that include mean absolute error (MAE), coefficient of determination (R²), Block-R\textsuperscript{2}, root mean squared error (RMSE), mean absolute percent error (MAPE), mean error (ME), and edge error metric (EE) (see Appendix for error metric equations). These metrics were applied to VibrantVS and the baseline models after masking lidar CHM pixels $<$ 2 meters in order to only include pixels that have a high potential to represent trees or higher vegetation. This reduced noise from low-lying vegetation or surface features that are not relevant for our canopy height study \citep{ferraz2012}. We also masked out lidar pixels where there were no data due to incomplete test tiles that may border flight path edges.  
To apply our error metrics, we spatially intersected all lidar CHM test tiles (500 $\times$ 500-meter$^2$ area; 1,000 $\times$ 1,000 pixels at 0.5-meter spatial resolution) with VibrantVS and the baseline model CHMs. We then calculated the pixel-wise error metric across all 1,000,000 pixels in the test tile, accounting for all masked pixels. We repeated this for all test tiles and took the median error metric across tiles aggregated by:
\begin{enumerate}
    \item The majority spatial intersection of tiles with the corresponding EPA L3 ecoregion to determine ecoregion-level accuracies.
    \item Individual height bins within each tile to determine height-class accuracies.
\end{enumerate}

\section{Results}

VibrantVS outperforms all baseline models with an overall median MAE of 2.71 meters, Meta: 4.83 meters; LANDFIRE: 5.96 meters; and ETH: 7.05 meters (Table \ref{tab:baseline_summary}). The overall mean error results show a median underestimation of canopy heights of 1.11 meters by VibrantVS, 4.03 meters by Meta, and a median overestimation by LANDFIRE of 0.92 meters and by the ETH model of 5.65 meters (Table \ref{tab:baseline_summary}).

\begin{figure*}[htbp!]
    \centering
    \includegraphics[width=\textwidth]{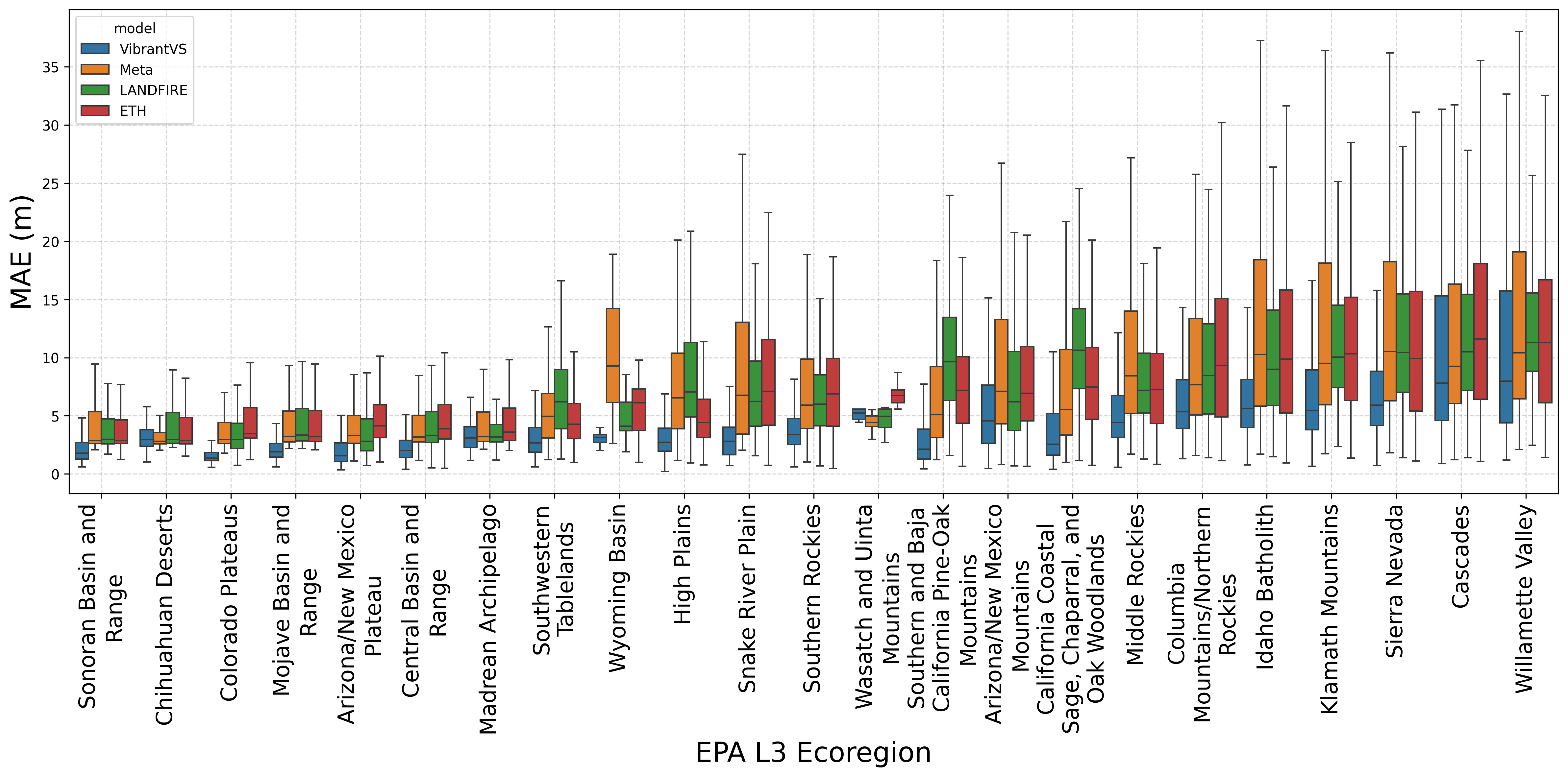}  
    \caption{Box and whisker plots of test tile level MAE by model and ecoregion sorted by lowest to highest median MAE across all models.}
    \label{fig:mae_by_ecoregion}  
\end{figure*}

Furthermore, the VibrantVS model consistently demonstrates lower error ranges in nearly all ecoregions analyzed, as evidenced by the narrower box plot widths for this model (Fig. \ref{fig:mae_by_ecoregion}). This highlights the higher CHM accuracy of VibrantVS at the tile level.

\begin{figure*}[htbp!]
    \centering
    \begin{subfigure}[t!]{0.46\textwidth}
        \centering
        \includegraphics[width=\textwidth]{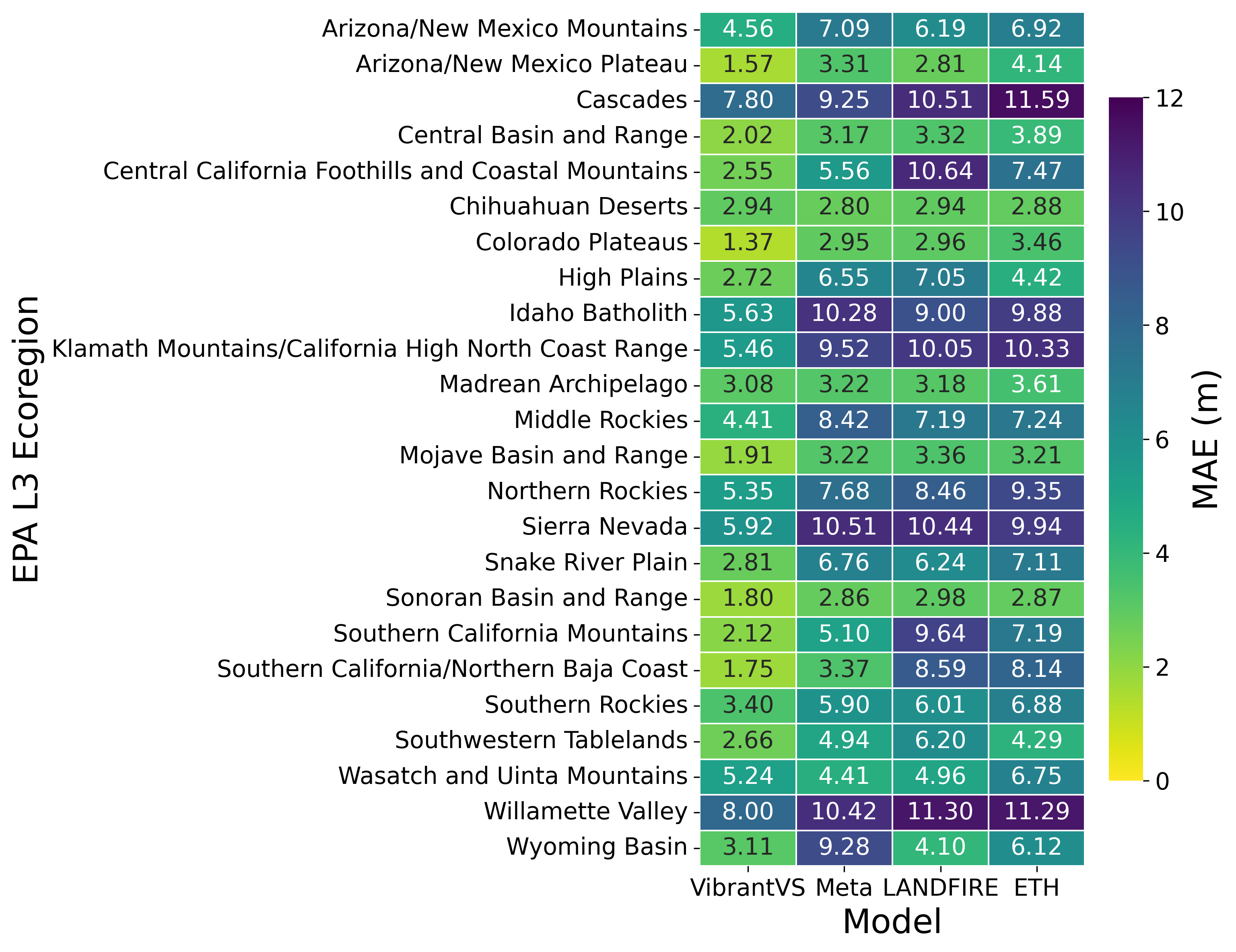}
        \caption{}
    \end{subfigure}
    \hfill
    \begin{subfigure}[t!]{0.46\textwidth}
        \centering
        \includegraphics[width=\textwidth]{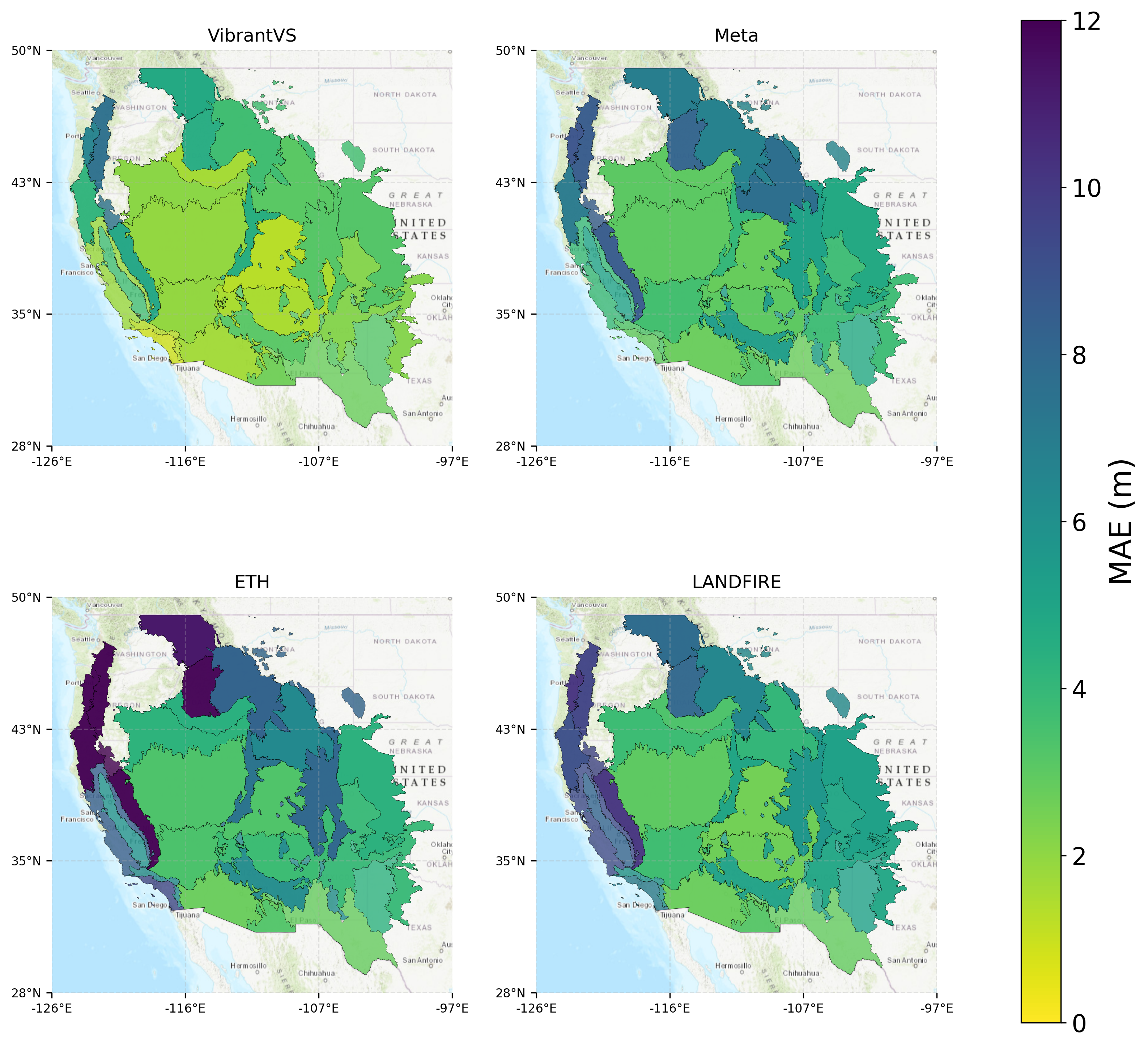}
        \caption{}
    \end{subfigure}
    \caption{(a) Results of MAE by ecoregion within all test tiles. (b) Map of MAE by ecoregions within all validation and test tiles to give complete display of the spatial relationship of baseline model performance.}
    \label{fig:mae_by_ecoregion_table_and_map}
\end{figure*}

The observed error of VibrantVS is consistently the lowest in desert and low-elevation ecoregions, where vegetation heights are typically below 10 meters in height, demonstrating high accuracy in shrub and grass systems. As shown in Figure \ref{fig:chm_bins_visual}, the error increases substantially in western forests where the 95th percentile of lidar heights is above 25 meters, suggesting taller long-lived trees such as coastal redwoods and cedars pose a substantial challenge for all baseline models and the VibrantVS model (Fig. \ref{fig:chm_bins_visual}, \ref{fig:mae_by_ecoregion}). This is reflected in the MAE map (Fig. \ref{fig:mae_by_ecoregion_table_and_map}b) where MAE increases within ecoregions where there are larger species of western trees (see Fig. \ref{app:lidar_boxplots} for canopy height box plots by ecoregion).

\begin{figure}[htbp!]
    \centering
    \includegraphics[width=0.5\textwidth]{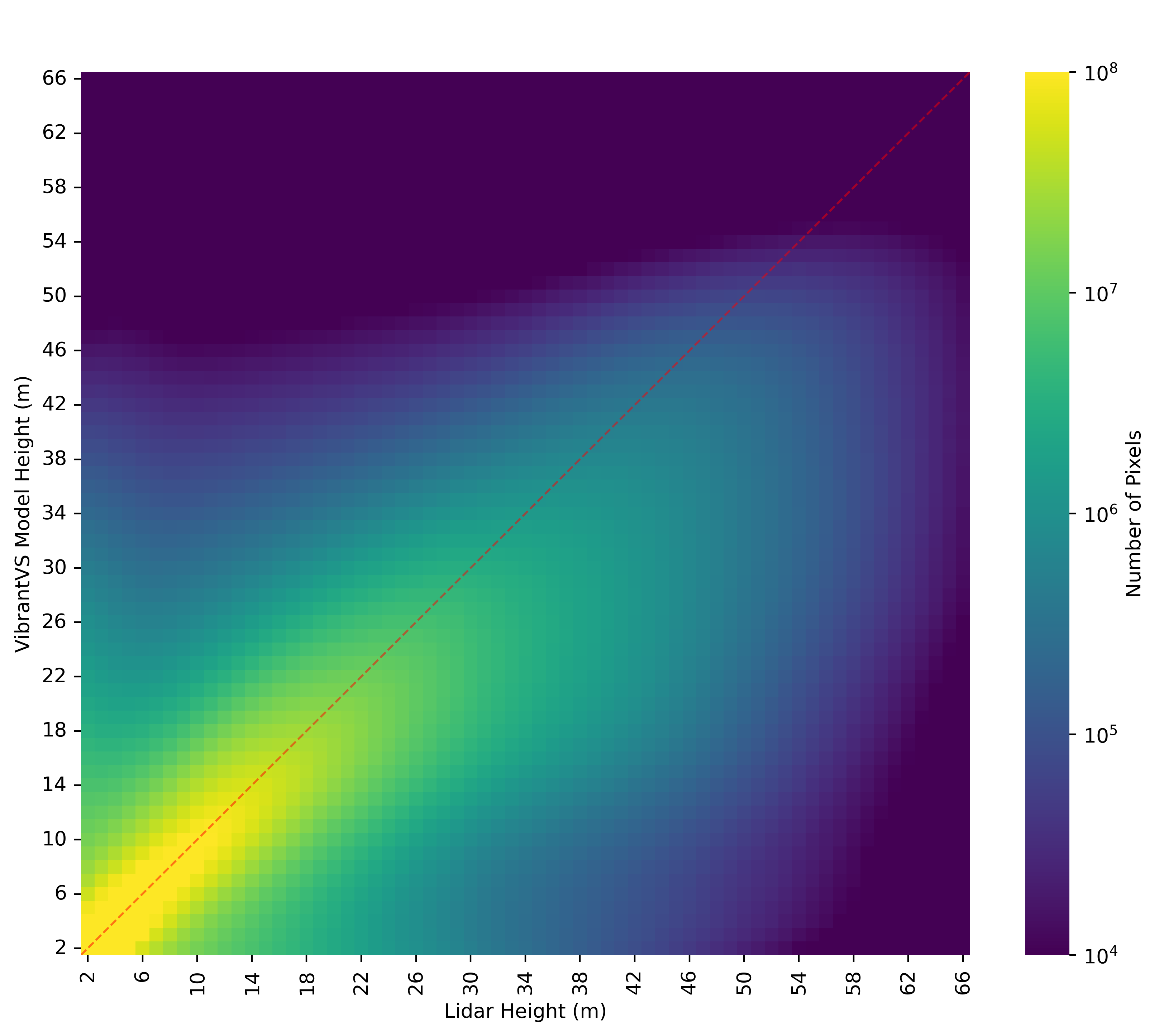}  
    \caption{2D histogram plot of VibrantVS canopy heights vs lidar-derived canopy heights across test samples.}
    \label{fig:model_hist2d}  
\end{figure}

Finally, our evaluation shows that all four models tend to truncate tree heights above 50 meters, with ETH's model performing best in this height range (Fig. \ref{fig:chm_bins_visual}, \ref{fig:model_hist2d}). 

\begin{figure}[htbp!]
    \centering
    
    \includegraphics[width=0.49\textwidth]{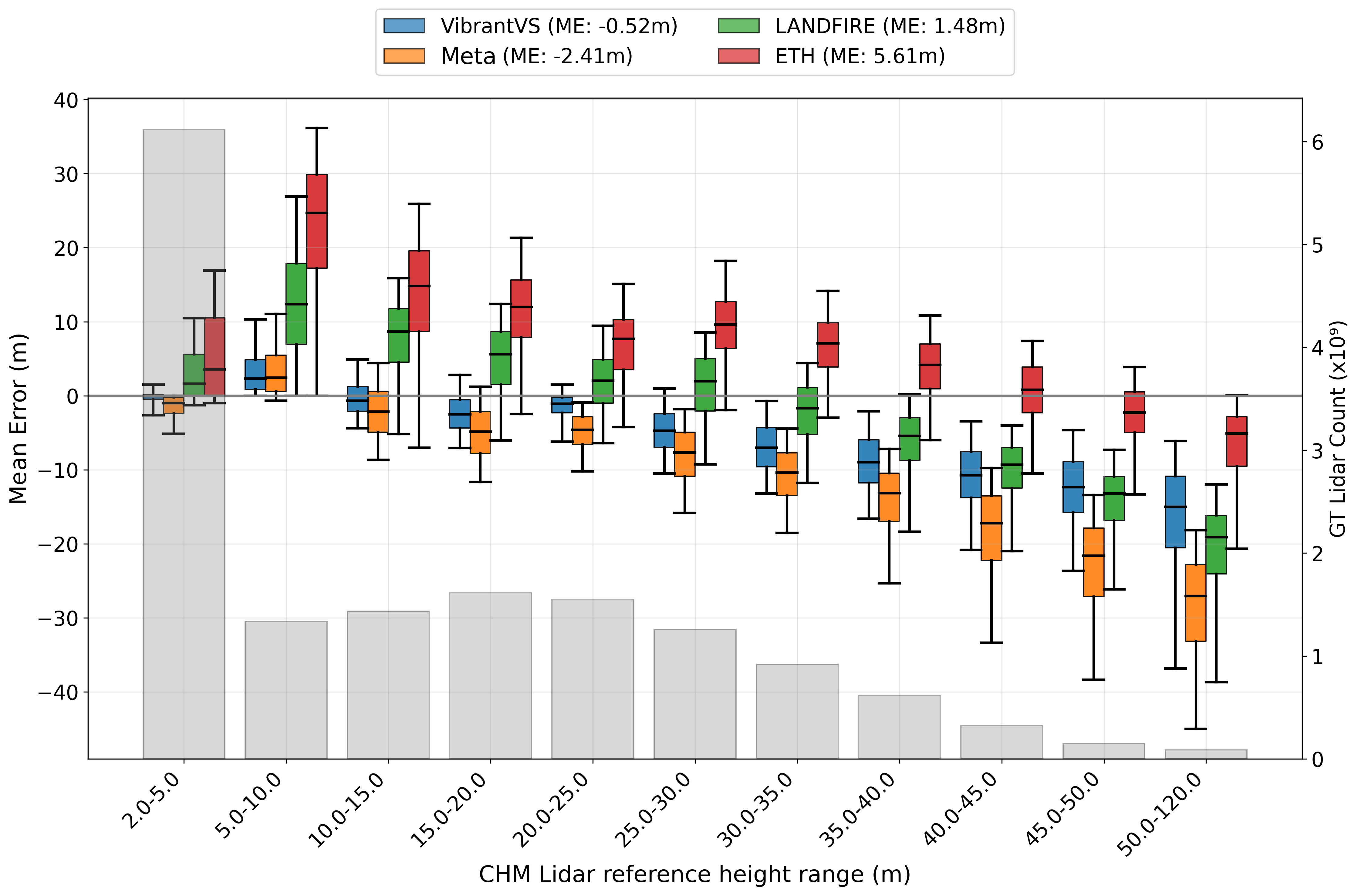}  
    \caption{Box plots of mean error in different height bins and histogram of sample counts per height bin.}
    \label{fig:mean_error_by_height_bin}  
\end{figure}

The mean-error results by canopy height bins show that VibrantVS has less bias than the other baseline models in the range of 2 to 25 meters. These height classes coincide with the majority of the lidar pixels in our test data. Beyond these heights, LANDFIRE and ETH tend to outperform VibrantVS when compared to lidar (Fig. \ref{fig:mean_error_by_height_bin}). However, while the VibrantVS and Meta models tend to show increasingly negative mean error at higher height bins, both LANDFIRE and ETH strongly overestimate lower canopy height bins. 

\begin{figure*}[htbp!]
    \centering    
    \includegraphics[width=\textwidth]{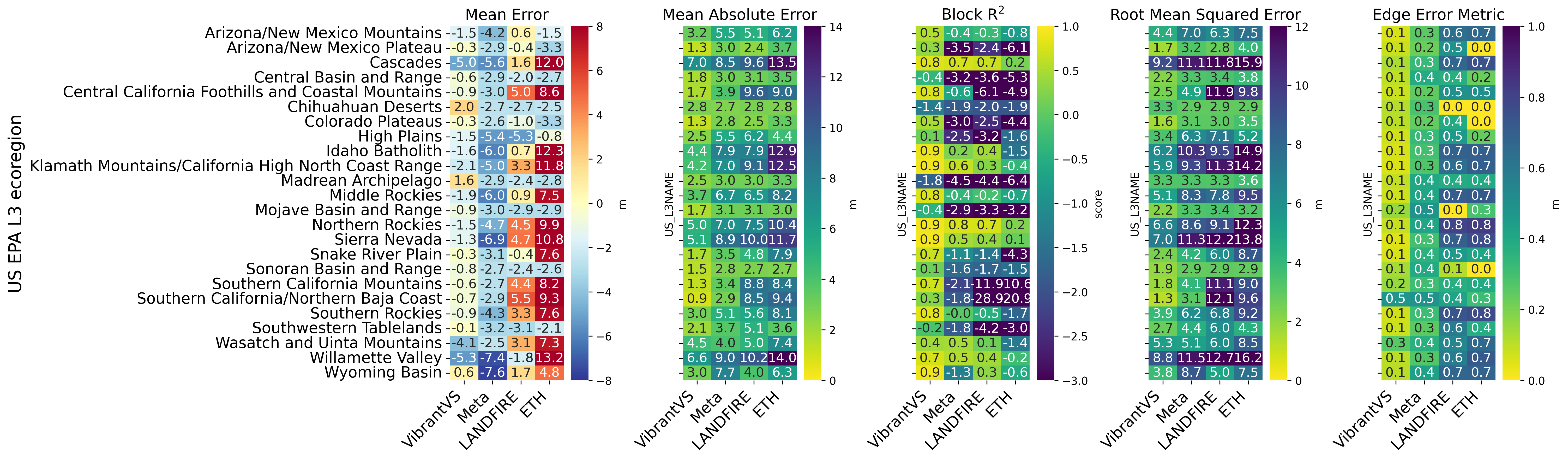}  
    \caption{Comparison of model performance by ecoregion for every error metric.}
    \label{fig:by_ecoregion_metrics}  
\end{figure*}

\begin{table}[ht]
    \centering
    \caption{Summary of baseline model across all test tiles for all error metrics.}
    \begin{tabular}{l|r|r|r|r}
    
    Model Name & MAE & Mean Error & Block-R$^2$ & Edge Error \\
    \hline

    VibrantVS & 2.71 & -1.11 &  0.69 & 0.08 \\
    Meta & 4.83 & -4.03 & -0.60 & 0.30 \\
    LANDFIRE & 5.96 & 0.92 & -1.45 & 0.63 \\
    ETH & 7.05 & 5.65 & -1.85 & 0.64 \\

    \label{tab:baseline_summary}
    \end{tabular}
\end{table}
The summaries in Table \ref{tab:baseline_summary} indicate that VibrantVS outperforms the other baseline models for mean absolute error (MAE), Block-R$^2$, and edge error metrics, suggesting a better ability to synthesize lidar-like CHM outputs.
The median Block-R$^2$ \citep{tolan2024very} numbers across all tiles also reflect the performance trend of MAE and ME (compare Table \ref{tab:baseline_summary} and Fig. \ref{fig:by_ecoregion_metrics}). 

\begin{figure*}[ht]
    \centering
    \includegraphics[width=\textwidth]{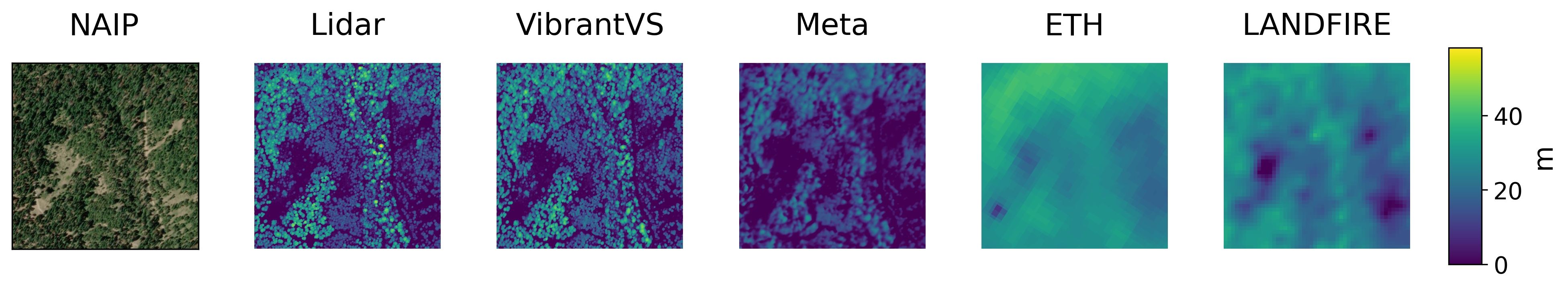}
    \caption{Qualitative evaluation of the various baseline models compared to the original lidar data.}
    \label{fig:qualitative_plot}
\end{figure*}

\subsection{Qualitative analysis}
\begin{figure*}[ht!]
    \centering
    \includegraphics[width=\textwidth]{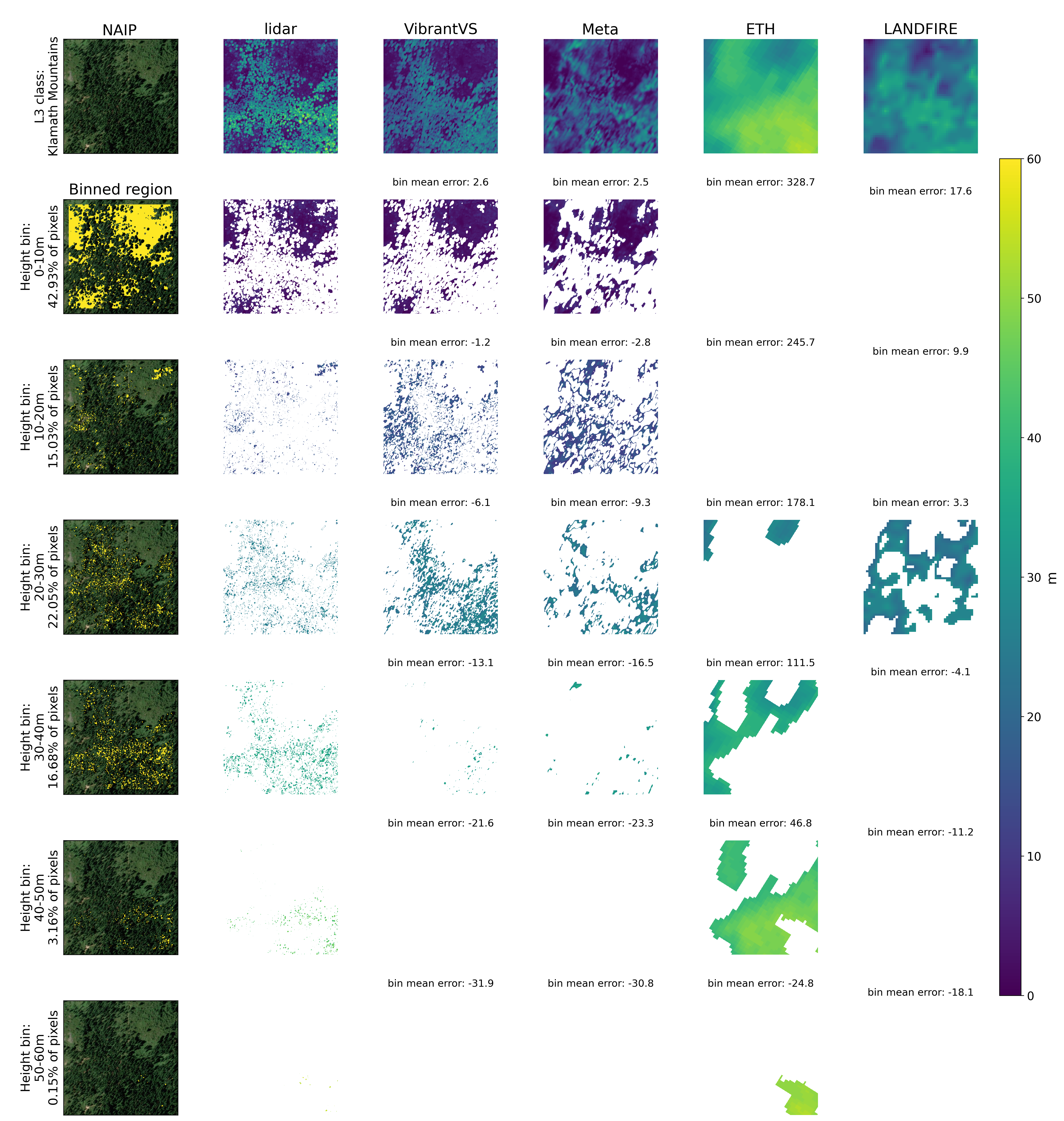}
    \caption{Comparison of model performance at different height class bins. }
    \label{fig:chm_bins_visual}
\end{figure*}

\begin{figure}[htbp!]
    \centering    
    \includegraphics[width=0.49\textwidth]{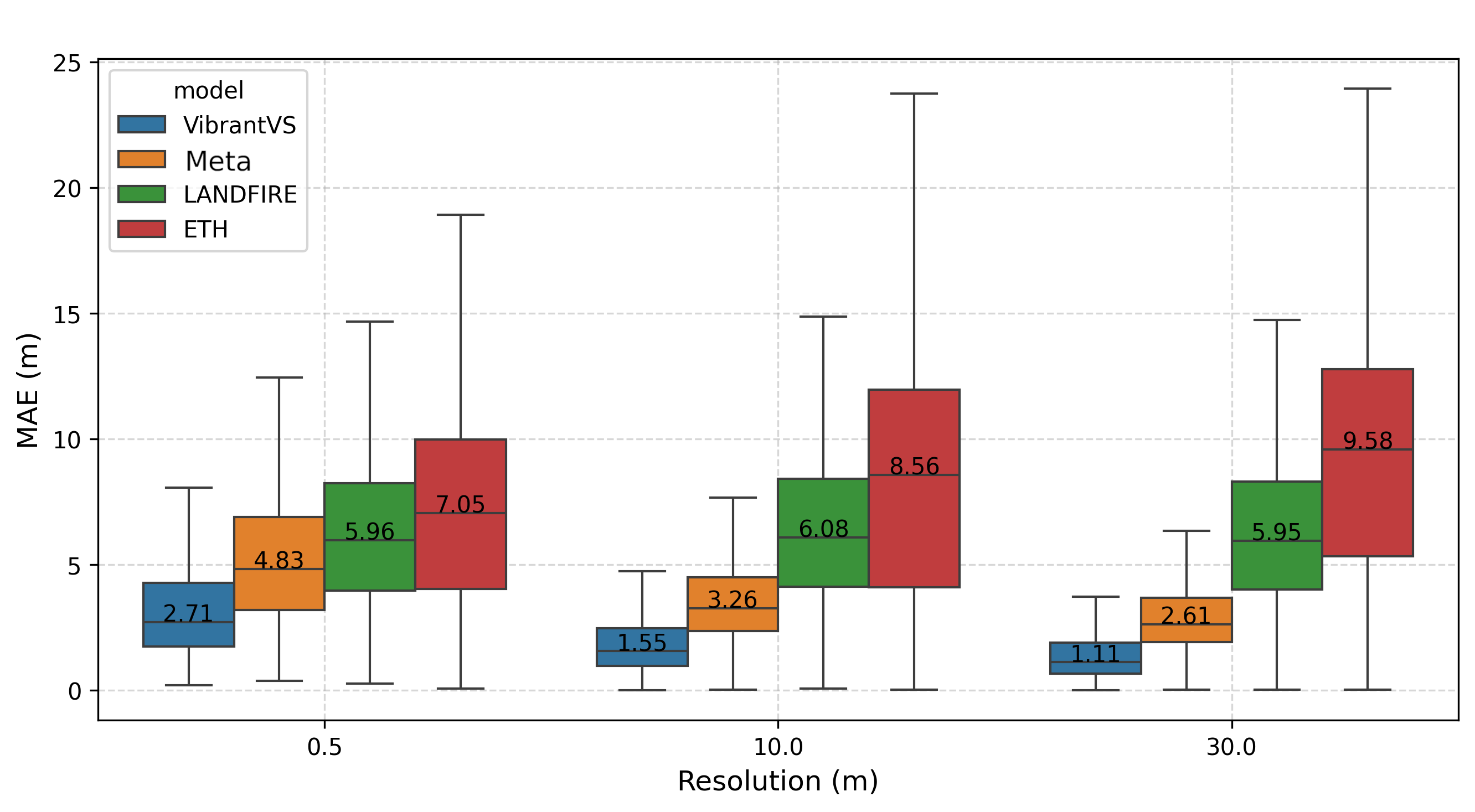}  
    \caption{Comparison of model performance at varying resolutions with the target lidar data resampled with averaging method. Median MAE values annotated in box whisker plots.}
    \label{fig:resolution_comparison}  
\end{figure}

The Meta and VibrantVS CHM estimates provide finer resolution data that are visually closer to the corresponding aerial lidar data, resulting in higher fidelity for fine-scale applications (Fig. \ref{fig:qualitative_plot}). This is made evident by the lower median edge errors (EE) observed for these models compared to ETH and LANDFIRE (see Appendix Eq. \ref{eq:ee}) between the models, where VibrantVS's EE equals 0.08, compared to Meta with 0.3, and ETH and LANDFIRE at 0.63 and 0.64 (Table \ref{tab:baseline_summary}). Considering the coarser resolution of the LANDFIRE and ETH data, their estimates often result in a constant value and a general overestimation of the CHM values compared to lidar and a lower underestimation in canopy heights beyond 30 meters (Fig. \ref{fig:chm_bins_visual}). When we aggregate VibrantVS's results to 10 and 30 meters and compare them with the baseline models at the same resolution, we find that the trends in differences between the overall median MAEs are consistent with the 0.5-meter results. (Fig.\ref{fig:resolution_comparison}).

\section{Discussion}
In this paper, we evaluate a novel vision transformer model against other baseline canopy height models. Our results demonstrate that the VibrantVS model produces a high-accuracy, high-precision CHM across ecoregions in the western United States. 

The overall performance of the VibrantVS model is likely better than the baseline models because we use a larger quantity of training data from a broader set of ecoregions (Table \ref{tab:lidar_summary}, Fig. \ref{fig:study_area_counts}). Our training and test data are based on a CHM created from aerial lidar from a wide range of ecoregions in the western United States with a total surface area of \~190,000 km$^2$, allowing our model to learn a wider range of landscape canopy-height patterns. In contrast, the baseline models were trained on more limited regions: NEON sites, Forest Inventory and Analysis (FIA) field plots, or select GEDI plots, that may represent sets of conditions that are too limited to create a robust model. Although other factors such as the use of higher resolution NAIP imagery have contributed as well, we attribute most of our performance gains to our robust training dataset and especially to our selection of model improvements for our ViT. \cite{WAGNER2024114099} have used a U-Net-based approach with the same type of NAIP input data to estimate canopy heights, but evaluated the performance only in California and therefore we could not include them in this analysis.  
 
The temporal alignment of our NAIP images is within a single year of the acquisition date of the aerial lidar for the data used to train VibrantVS. However, this is not the case for the other baseline models. They all aim to represent a CHM from the year 2020; however, their source data come from a wider range of acquisition years: Maxar mosaic data from 2018 to 2022 \citep{tolan2024very}, Landsat data from 2016 \citep{LANDFIREFuels2022}, or Sentinel-2-L2A data from 2020 \citep{lang2023high}. The temporal discrepancy with the acquisition years from 2016 to 2021 of the aerial lidar led to larger estimation errors. However, even when we only compare the models with corresponding lidar data from 2019 to 2021, the trend in the difference in MAE numbers is maintained (Fig. \ref{app:lidar_year_comparison}). This highlights a specific advantage of the use of NAIP imagery as a model input, given NAIP's three-year-or-less temporal return interval. This cadence across states allows for monitoring of canopy heights following major disturbances and can provide updated information of the forest structure without the cost of lidar acquisition.   

VibrantVS utilizes higher resolution source image data from NAIP compared to the other models. A resolution of 0.5 meters provides better granularity of the features within the image and allows us to provide native estimates at 0.5 meters, which aligns with the aerial lidar CHM that we compare against. Meta provides a CHM at 1-meter resolution, ETH at 10 meters, and LANDFIRE at 30 meters. Given our use of nearest-neighbor resampling to compare our baselines versus lidar, we expected high-resolution comparisons to inflate MAE for baseline models due to their lower native resolutions compared to VibrantVS, but this is not the case when observing MAE for aggregations at spatial resolutions at 10 and 30 meters. When we compare all models at 10 and 30-meter resolutions to lidar CHM at these same resolutions, the trend in differences of overall median MAE is maintained relative to the 0.5-meter results (Fig. \ref{fig:resolution_comparison}).

The coarser resolution of the ETH and LANDFIRE models restricts their use to coarse-level applications, for which the trends in overestimation and underestimation at different tree heights could be accounted for with correction factors by height bins. The high-resolution of the VibrantVS model provides the opportunity to generate downstream forest structure products such as Tree Approximate Object (TAO) segmentation \citep{Jeronimo_2018}. The spatial resolution of a CHM is a critical factor in accurately segmenting individual trees, which often serve as inputs for various forest management activities. Higher resolutions, such as 0.5 meters, can provide detailed canopy representations that a 1-meter pixel resolution model might not capture, leading to less precise segmentation \citep{lidRbook2024}. Additionally, the CHM rasters can serve as input to wildfire spread models, allowing practitioners to better capture discontinuities in fuels that may slow or stop fire progression \citep{ritu_2021}. 

From our analysis we determine that the following improvements can be made in future research:
\begin{enumerate}
    \item More refinements can be made to the evaluations of the lowest height bin. MAE values are higher in this category because the spatial footprint of tree crowns appear larger in NAIP data compared to lidar-based tree crowns. This aerial image-based effect causes the VibrantVS model to infer a wider canopy crown that results in an overestimation of values where the lidar data are closer to 0. If we use the block metrics of Tolan et al. \citep{tolan2024very}, rather than the pixel-wise metrics of Lang et al. \citep{lang2023high}, then we expect the lowest bin error of overestimation to disappear. 
    \item More importantly, we have to address the underestimation of height among trees that are taller than 35 meters, and especially trees taller than 50 meters. There is an imbalance in the number of training samples of tall trees.  
    Trees taller than 50 meters represent less than 0.5\% of our training data. We are undertaking a specific retraining exercise to address this underestimation and to compensate for the label imbalance. 
    
    \item We would like to integrate topography data into our model, as we suspect that this could improve accuracy, especially on steep slopes.
    \item We are also planning to expand our range of applications to more ecoregions, especially in the central and eastern United States. This would allow us to include additional tree species and also shrub vegetation types and to evaluate CHMs at heights lower than 2 meters. 
\end{enumerate}

\section{Conclusion}
This paper presents a comprehensive analysis of three baseline CHM models and our custom vision transformer model (VibrantVS), which is based on NAIP imagery. Our findings indicate that, while each model has its strengths, their success varies significantly between different ecoregions. VibrantVS outperforms all three baseline models in both accuracy and precision across the majority of ecoregions and ranges of canopy heights. It comes with the distinct advantage of being able to produce a very high resolution (0.5 meters) CHM and can be updated with a cadence of three years or less. Additionally, VibrantVS is based on publicly and freely available data in the United States and can be an alternative to collecting aerial lidar at high costs. 

The very high-resolution CHMs from VibrantVS can be used in the TAO segmentation process, which allows for the downstream estimation of key forest-structure variables, such as trees per acre, basal area, timber volume, canopy base height, and biomass. These variables are essential for decision making in forestry and land management. 
These derived variables can also be integrated into existing fire models to predict fire risk and behavior at high resolution.

\section*{Acknowledgments}
The authors would like to thank Leo Tsourides and Derek Young for their support and consulting in this research. Additional gratitude is extended to Ian Reese, Fabian Döweler, Finlay Thompson, and the rest of the Dragonfly Data Science team for initial proof-of-concept work. The authors thank NASA and USDA for their support. Luke J. Zachmann was funded in part by USDA grant number NR233A750004G042, subaward number 5; Tony Chang, Andreas Gros, and Vincent A. Landau were funded in part by NASA grant number 80NSSC22K1734.

% \bibliographystyle{cas-model2-names}
\bibliographystyle{abbrvnat}
\bibliography{references}
\newpage
\appendix
\section{Appendix}
\renewcommand{\thefigure}{A\arabic{figure}}
\setcounter{figure}{0}
\renewcommand{\thetable}{A\arabic{table}}
\setcounter{table}{0}
\subsection{Additional Tables and Figures}
\begin{figure*}[htbp!]
    \centering
    \includegraphics[width=\textwidth]{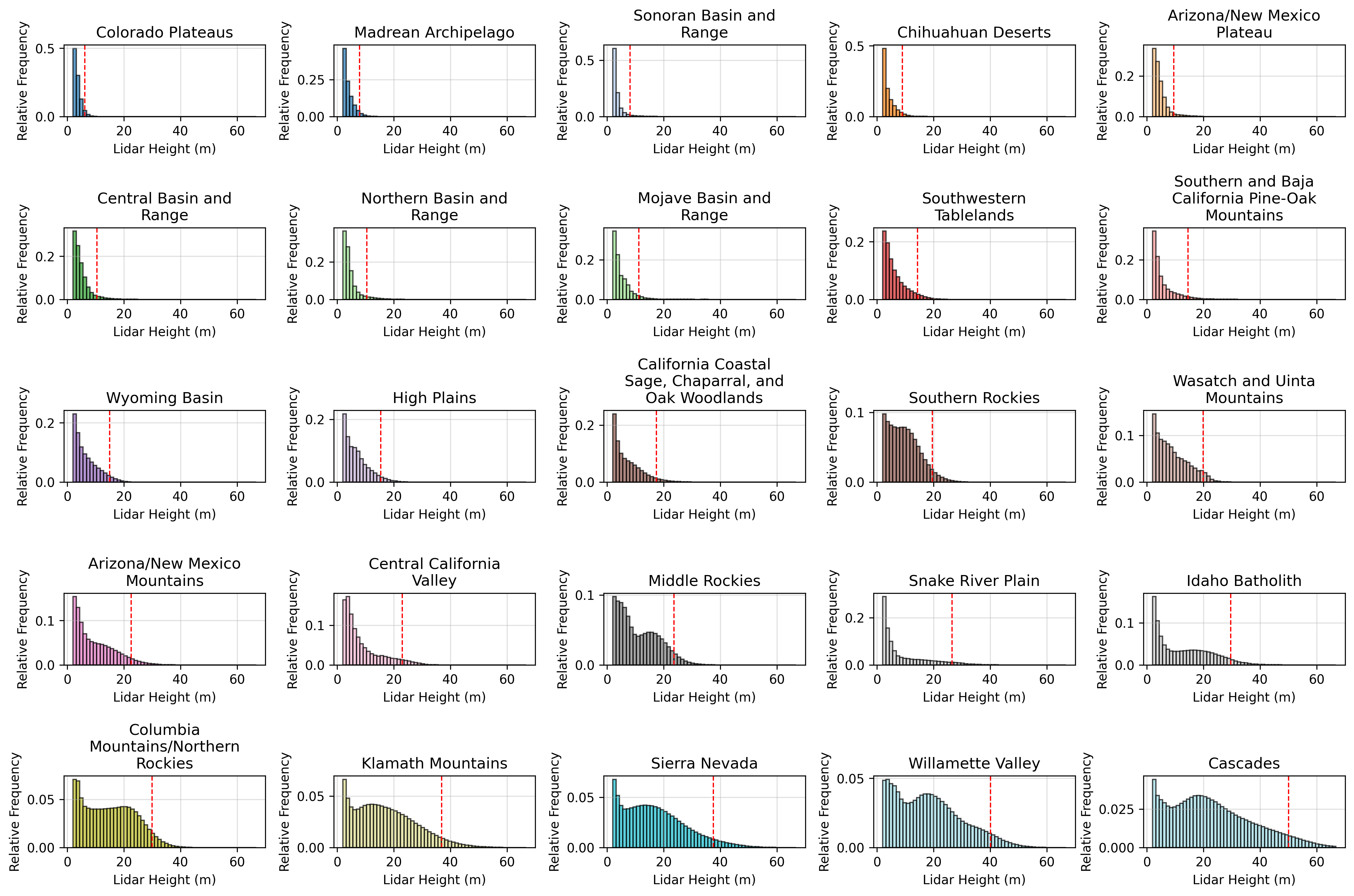}  
    \caption{Lidar sample pixel-wise height distributions within each of the EPA Level 3 ecoregions. Red dashed line represents the 95th percentile of the height distribution.}
    \label{app:ecoregion_histograms}  
\end{figure*}

\begin{table*}[htbp!]
\centering
\begin{tabular}{p{1.75cm}|p{1.25cm}|p{1.25cm}|p{1.25cm}|p{1.9cm}|p{2cm}|p{1.6cm}}
\hline
\textbf{Model} & \textbf{Spatial Resolution} & \textbf{Spatial Extent}&\textbf{Tempor. Coverage}&\textbf{Model Type} & 
\textbf{Predictor Data} & \textbf{Label Data} \\ 
\hline
VibrantVS & 0.5-meter & Western United States& 2014-2021& Multi-task vision transformer & 4-band NAIP imagery & USGS 3DEP aerial lidar \\  \hline
Meta \citep{tolan2024very} & 1-meter & Global & 2020 & Encoder, dense prediction transformer, correction and rescaling network &  Maxar Vivid2 0.5-meter resolution mosaics &  NEON aerial lidar CHMs, GEDI data, and a labeled 9000 tile tree/no tree segmentation dataset \\ \hline
LAND FIRE \citep{rollins2009landfire} & 30-meter & United States& 2016, 2020, 2022, 2023&  Regression-tree based methods & Spectral information from Landsat, landscape features such as topography, and biophysical information &  Field-measured height \\ \hline
ETH \citep{lang2023high} & 10-meter & Global&2020&  Deep learning ensemble &  Sentinel-2-L2A multi-spectral imagery, sin-cos embeddings of longitudinal coordinates &  Sparse GEDI lidar data \\ 
\hline
\end{tabular}
\caption{Summary of CHM models, spatial resolutions, predictor data, and label data.}
\label{app:chm_models}
\end{table*}

\begin{figure*}[htbp!]
    \centering    
    \includegraphics[width=\textwidth]{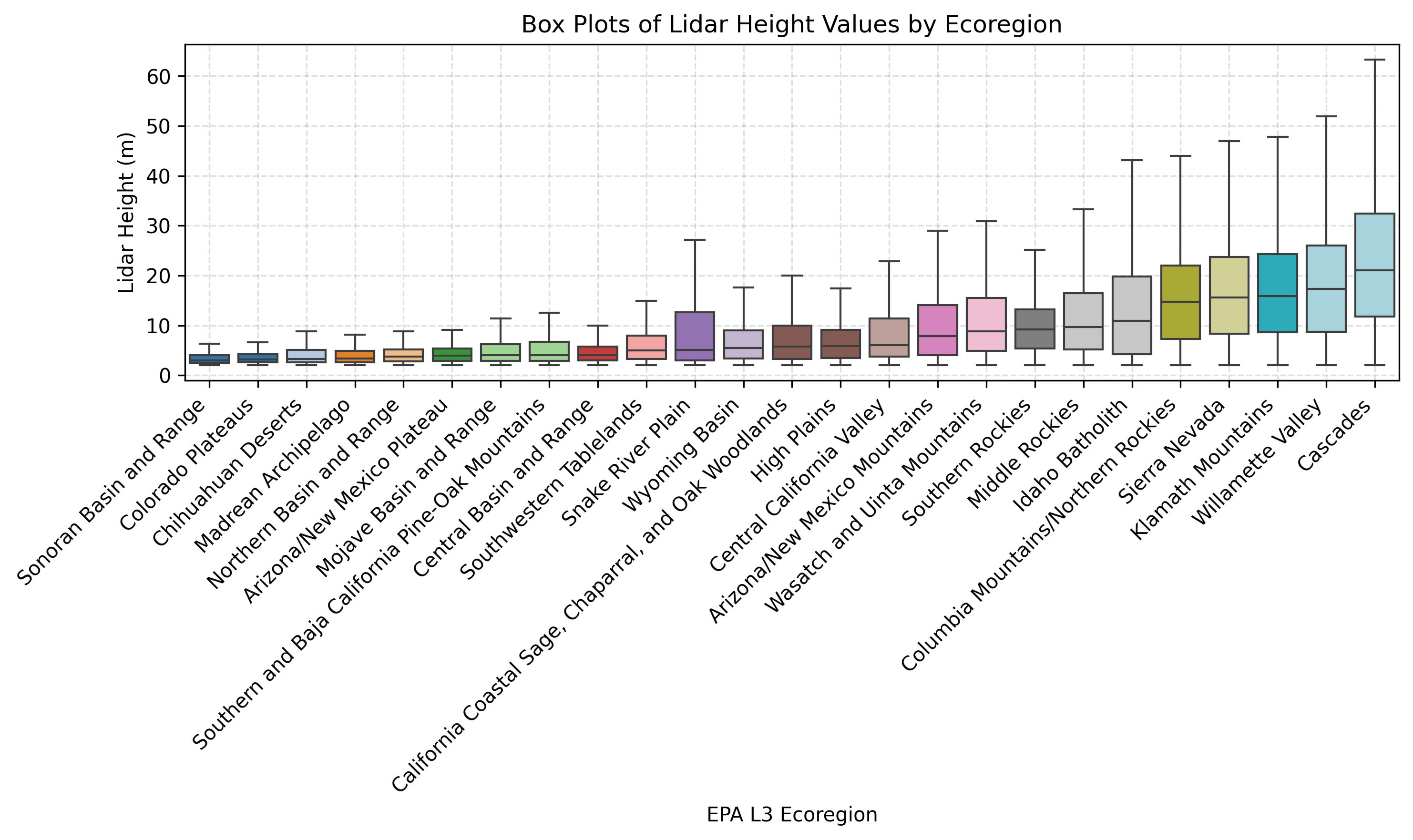}  
    \caption{Lidar sample tile height box plots within each EPA Level 3 ecoregion.}
    \label{app:lidar_boxplots}  
\end{figure*}

\begin{figure*}[htbp!]
    \centering    
    \includegraphics[width=\textwidth]{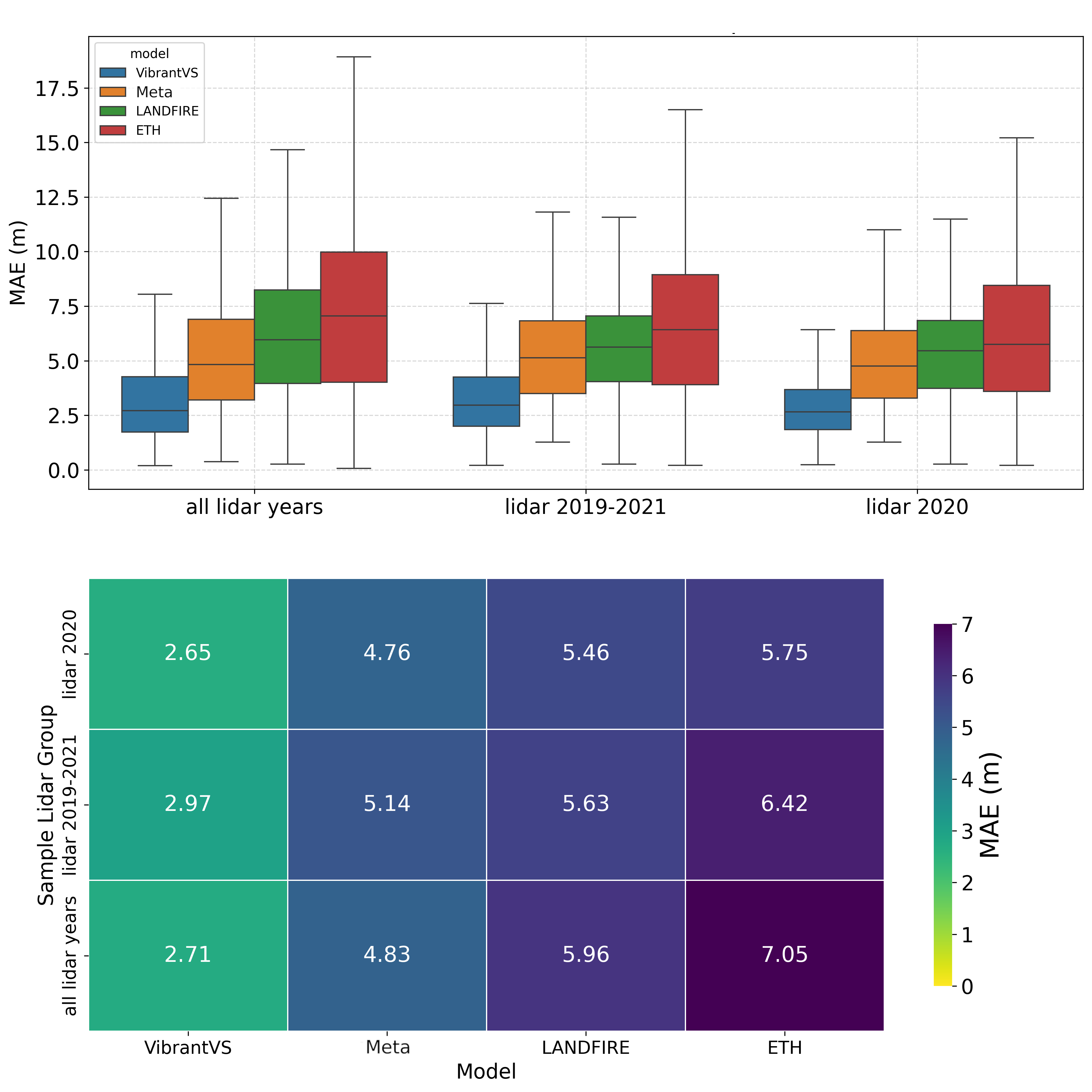}  
    \caption{Comparison of model performance at different year groups of lidar acquisition determine temporal mismatch error.}
    \label{app:lidar_year_comparison}  
\end{figure*}

\subsection{Error metrics}
\begin{enumerate}
    \item Mean Absolute Error (MAE)
    \[
    \mathrm{MAE} = \frac{1}{n} \sum_{i=1}^n \left| y_i - \hat{y}_i \right|
    \]

    \item Block-R\textsuperscript{2}
    \[
    R^2_{\text{block}} = 1 - \frac{\sum_{b=1}^B \sum_{i \in b} (y_i - \hat{y}_i)^2}{\sum_{b=1}^B \sum_{i \in b} (y_i - \bar{y}_b)^2}
    \]
    Where \( B \) is the number of blocks, \( y_b \) is the ground truth value in block \( b \), \( \hat{y}_b \) the model estimate for block \( b \), and \( \bar{y}_b \) is the mean of the ground-truth values in block \( b \). 

    \item Root Mean Square Error (RMSE)
    \[
    \mathrm{RMSE} = \sqrt{ \frac{1}{n} \sum_{i=1}^n \left( y_i - \hat{y}_i \right)^2 }
    \]
    
    \item Mean Error (ME)
    \[
    \mathrm{ME} = \frac{1}{n} \sum_{i=1}^n \left( \hat{y}_i - y_i \right)
    \]
    \item Edge Error Metric (EE) \label{eq:ee}
    \[
    \mathrm{EE} = \frac{1}{n} \sum_{i=1}^n \left| E(\hat{y}_i) - E(y_i) \right|
    \]
    Where \( E(\cdot) \) represents Sobel edge detection operation on the data (also compare Tolan et al. (\citeyear{tolan2024very})).
\end{enumerate}

\end{document}